\documentclass[11pt,a4paper]{article}

\usepackage[english]{babel}
\usepackage[utf8]{inputenc} 
\usepackage{fontenc}        
\usepackage[natbibapa]{apacite}
\usepackage{lscape}  
\usepackage{rotating} 

\usepackage{footnote}

\usepackage[table,xcdraw]{xcolor} 
\usepackage{booktabs}       
\usepackage{multirow}       
\usepackage{nicefrac}       
\usepackage{enumitem}       
\usepackage[colorinlistoftodos]{todonotes} 
\usepackage{amsmath}        
\usepackage{amssymb}        
\usepackage{amsfonts}       
\usepackage{graphicx}       
\usepackage{subcaption}     
\usepackage{appendix}       
\usepackage{soul}           
\usepackage{microtype}      
\usepackage{textcomp}       
\usepackage{url}            
\usepackage[colorlinks=false, linkcolor=blue, citecolor=black]{hyperref}      

\begin{document}
\title{Comparison of State-of-the-Art Deep Learning APIs for Image Multi-Label Classification using Semantic Metrics}
\date{\vspace{-5ex}} 

\author{
Adam Kubany,\footnote{Corresponding author. 
\newline E-mail addresses: adamku@post.bgu.ac.il (A.Kubany), benishue@gmail.com (S.Ben Ishay), rubensac@post.bgu.ac.il (R.Ohayon), armin@bgu.ac.il (A.Shmilovici), liorrk@post.bgu.ac.il (L.Rokach), tomerdoi@post.bgu.ac.il (T.Doitshman)
\newline An earlier version of this article was entitled "Semantic Comparison of State-of-the-Art Deep Learning APIs for Image Multi-Label Classification."} \ Shimon Ben Ishay, Ruben Sacha Ohayon, \\Armin Shmilovici, Lior Rokach, Tomer Doitshman\\
Department of Software and Information System Engineering\\
Ben-Gurion University of the Negev, Israel\\
}

\maketitle
\begin{abstract}
Image understanding heavily relies on accurate multi-label classification. In recent years, deep learning algorithms have become very successful for such tasks, and various commercial and open-source APIs have been released for public use. However, these APIs are often trained on different datasets, which, besides affecting their performance, might pose a challenge to their performance evaluation. This challenge concerns the different object-class dictionaries of the APIs' training dataset and the benchmark dataset, in which the predicted labels are semantically similar to the benchmark labels but considered different simply because they have different wording in the dictionaries. To face this challenge, we propose semantic similarity metrics to obtain richer understating of the APIs predicted labels and thus their performance. In this study, we evaluate and compare the performance of 13 of the most prominent commercial and open-source APIs in a best-of-breed challenge on the Visual Genome and Open Images benchmark datasets. Our findings demonstrate that, while using traditional metrics, the Microsoft Computer Vision, Imagga, and IBM APIs performed better than others. However, applying semantic metrics also unveil the InceptionResNet-v2, Inception-v3, and ResNet50 APIs, which are trained only with the simple ImageNet dataset, as challengers for top semantic performers.

\end{abstract}

\providecommand{\keywords}[1]
{\small	\textbf{\textit{Keywords---}} #1}

\section{Introduction}
Accurate semantic identification of objects, concepts, and labels from images is one of the preliminary challenges in the quest for image understanding. The race to achieve accurate label classification has been fierce and became even more so as a result of public competitions such as the ImageNet Large Scale Visual Recognition Challenge (ILSVRC) \citep{Russakovsky2015ImageNetChallenge}, and the release of benchmark datasets such as the YFCC100M \citep{Thomee2016YFCC100M:Research}, Visual Genome \citep{Krishna2017VisualAnnotations}, MS-COCO \citep{RefWorks:202}, and Open Images \citep{Kuznetsova2020TheScale}. Different learning approaches for multi-label classification have been suggested to answer this call. Tsoumakas and Katakis \citep{RefWorks:247, RefWorks:249} divided these approaches into two main categories: 1) \textit{problem transformation methods} which transform the problem into one or more single-label classification problem and then aggregate the results into a multi-label representation; and 2) \textit{algorithm adaptation methods} which solve the multi-label prediction problem as a whole, directly from the data. In 2012, Madjarov et al. \citep{RefWorks:245} introduced a third category of methods, referred to as \textit{ensemble methods}, which combine several classifiers to solve the multi-label classification problem. In this approach, each of the base classifiers in the ensemble can belong to either the problem transformation or algorithm adaptation method category. 

As the research field of multi-label classification advances, more and more effective approaches have been suggested \citep{RefWorks:245, RefWorks:248}. In recent years, deep learning methods, such as convolutional neural networks (CNN), and their variations, have demonstrated excellent performance \citep{Thomason2014IntegratingWild, RefWorks:234, Tran2016RichWild, RefWorks:262, RefWorks:340, RefWorks:341, He2017MaskR-cnn, RefWorks:339, Ren2015FasterNetworks}. Some of the more salient approaches were published as open-source or as commercial APIs, such as from Imagga \citep{RefWorks:303}, IBM Watson \citep{IBM2020VisualRecognition}, Clarifai \citep{RefWorks:304}, Microsoft Computer-Vision \citep{RefWorks:305}, Wolfram Alpha \citep{Wolfram2020WolframProject}, Google Cloud Vision \citep{Google2020GoogleVision}, DeepDetect \citep{DeepDetect2020DeepDetect}, YOLO \citep{Redmon2016YouDetection}, MobileNet \citep{Howard2017Mobilenets:Applications}, Inception \citep{RefWorks:267}, ResNet \citep{RefWorks:338} and InceptionResNet \citep{Szegedy2017Inception-v4Learning}. With these recent publications, the need for a best-of-breed performance comparison has arisen. While some comparisons between multi-label classification methods have been performed in the past \citep{RefWorks:245, RefWorks:248}, none of them included both open-source and commercial APIs in such extensive manner. 

In this study, we address this need and evaluate the performance of 13 state-of-the-art deep learning approaches with well-established multi-label evaluation metrics \citep{RefWorks:249, RefWorks:252}. While these metrics evaluate the performance based on whether the predicted labels exist in the ground truth list, they do not consider the semantic similarity between them. With this oversight, a fair comparison between the various APIs becomes challenging, as each of them can be trained on different datasets, and therefore include different object class dictionaries. 
To comply with this challenge, we propose to use semantic variations of traditional evaluation metrics, the word mover’s distance metric (WMD) \citep{RefWorks:315}, and state-of-the-art text embedding methods such as BERT \citep{Devlin2018Bert:Understanding}, RoBERTa \citep{Liu2019Roberta:Approach}, and XLNet \citep{Yang2019Xlnet:Understanding} as more insightful evaluation metrics. To the best of our knowledge, this study provides the most thorough evaluation of state-of-the-art deep learning multi-label image classification from both commercial and open-source APIs, and the only study to include semantic evaluation metrics.

The novel contributions of this study\footnote{The the APIs inference scripts and metrics applied in this study are available in \url{https://github.com/Adamkubany/Multilabel_Semantic_API_comparison}} are 1) demonstrating the significance of the proposed semantic similarity metrics to the APIs’ performance evaluation in particular when trained with different object class dictionaries, and 2) an extensive comparison of the predictive performance of 13 of the most prominent commercial and open-source publicly available APIs for multi-label image classification.

\section{Multi-Label Image Classification APIs}
We divide the classification APIs into two categories, commercial and open-source (Table \ref{tbl:APIs settings}). While the open-source APIs publish their network architecture, training schemes, and even make pre-trained models available for free use, the constantly improving commercial APIs do not reveal much about their proprietary algorithm, other than mentioning that they are based on deep neural networks. \\
\textbf{Commercial services:} These APIs are provided by various companies such as Imagga \citep{RefWorks:303}, IBM Watson (Visual Recognition) \citep{IBM2020VisualRecognition}, Clarifai \citep{RefWorks:304}, Microsoft (Computer Vision) \citep{RefWorks:305}, Wolfram Alpha (Image Identification) \citep{Wolfram2020WolframProject}, and Google (Cloud Vision) \citep{Google2020GoogleVision}. Among those, only the Microsoft’s Computer Vision API hint that it is based on a deep residual network (ResNet) \citep{RefWorks:338}, which has shown high performance in the past, nevertheless, they do not reveal the network size, applied training data or any other specific variations.  

We also include several top performing open-source frameworks with the capability of multi-label classification.\\
\textbf{DeepDetect:} The DeepDetect approach \citep{DeepDetect2020DeepDetect} is based on the GoogLeNet architecture with 22 layers network \citep{RefWorks:263} also known as the Inception-v1 network. Here, we evaluate the model provided by the Caffe framework which is pre-trained on the ImageNet dataset. \\
\textbf{VGG19} The very deep CNN also know as the "VGG" network \citep{Simonyan2014VeryRecognition}, is consisted of 16-19 CNN layers. Here we included the 19 layer version trained on the ImageNet dataset \citep{2020KerasApplications}.\\
\textbf{Inception v3:} The Inception-v3 approach \citep{RefWorks:267} implements variations of the inception-v1 blocks for accuracy optimization. We evaluated the Inception-v3 Keras implementation trained on the ImageNet dataset \citep{2020KerasApplications}.\\
\textbf{InceptionResNet v2:} The InceptionResNet-v2 approach includes the Inception-v4 advances together with residual connections \citep{Szegedy2017Inception-v4Learning}. Here, we used the ImageNet pre-trained model available via Keras implementation \citep{2020KerasApplications}.\\
\textbf{ResNet50:} The Residual network (ResNet) approach implements residual connections along traditional CNN. The ResNet offers various layer depths (50, 101, 152), here, we evaluated the very popular 50 layers network trained on the ImageNet dataset \citep{2020KerasApplications} and on the COCO dataset \citep{RefWorks:202, Olafenwa2020ImageAIDetection} as a performance reference.\\
\textbf{MobileNet v2:} This is the second version \citep{Sandler2018Mobilenetv2:Bottlenecks} of the MobileNet approach for mobile devices \citep{Howard2017Mobilenets:Applications}, it includes compact convolutional building blocks accompanied with residual ideas. We evaluated its ImageNet pre-trained model \citep{2020KerasApplications}.\\
\textbf{YOLO v3:} You Only Look Once (YOLO) third version approach \citep{Redmon2016YouDetection, Redmon2018Yolov3:Improvement} includes 106 convolution layers with residual connections and classify the image object in three scales. Here, we evaluated the DarkNet53 version pre-trained on the ImageNet dataset \citep{Redmon2020YOLOv3Classification} and on the COCO dataset \citep{Olafenwa2020ImageAIDetection} for additional performance reference.\\

\section{Evaluation Metrics}
Evaluating the different APIs' prediction performance requires standardized measures and metrics. Various metrics have been proposed in the past for such evaluation \citep{RefWorks:249, RefWorks:252}; these metrics can be divided into \textit{bipartition} and \textit{ranking} metrics \citep{RefWorks:249}. As none of the evaluated APIs provide a ranking for all of the labels in the ground truth dataset, we focus only on the bipartition metrics. For the metrics’ definitions let us denote $Y_i\in L={\left\{0,1\right\}}^q$ as the multi-label binary encoding label set of image \textit{i} from \textit{n} images dataset where \textit{L}  is the \textit{q} sized label set dictionary, and $Z_i$ as the multi-label binary encoding label set of image \textit{i} as predicted by the multi-label classifier \textit{h}; hence, $Z_i=h\left(x_i\right)\in L$, where $x_i\in\mathcal{X}$, is defined as the feature vector of image \textit{i}.
\subsection{Bipartition Metrics}
There are two types of bipartition evaluation metrics. Example-based bipartition evaluation metrics refer to various average differences of the predicted label set from the ground truth label set for all the dataset samples, whereas the label-based evaluation metrics evaluate each label separately and then average for all the labels.
\subsubsection{Example-Based}
The following \textit{Accuracy, Precision, Recall}, and $F_1$ metrics are standard metrics adapted for multi-label classification \citep{RefWorks:245, RefWorks:251}. 
Accuracy is defined as the Jaccard similarity between the predicted label set $Z_i$ and the ground truth label set $Y_i$, which is then averaged over all \textit{n} images.
\begin{equation}
    Accuracy=\frac{1}{n}\sum_{i=1}^{n}\frac{{|Y}_i\cap Z_i|}{|Y_i\cup Z_i|}	
\end{equation}
\textit{Precision} and \textit{Recall} are defined as the average proportion between the number of correctly predicted labels (${|Y}_i\cap Z_i|$) and either the number of predicted labels $Z_i$ or the number of ground truth labels $Y_i$. 
\begin{equation}
    Precision=\frac{1}{n}\sum_{i=1}^{n}\frac{{|Y}_i\cap Z_i|}{|Z_i|}	
\end{equation}
\begin{equation}
    Recall=\frac{1}{n}\sum_{i=1}^{n}\frac{{|Y}_i\cap Z_i|}{|Y_i|}	
\end{equation}
and $F_1$ is the harmonic mean between \textit{Precision} and \textit{Recall}.
\begin{equation}
    F_1=\frac{1}{n}\sum_{i=1}^{n}\frac{{2|Y}_i\cap Z_i|}{\left|Y_i\right|+|Z_i|}	
\end{equation}
\subsubsection{Label-Based}
Label-based metrics evaluate the performance of a classifier by first evaluating each label and then obtaining an average of all of the labels. Such averaging can be achieved by one of two conventional averaging operations, namely \textit{macro} and \textit{micro} averaging \citep{RefWorks:256}. For that purpose, any binary evaluation metric can be applied, but usually \textit{Precision}, \textit{Recall}, and their harmonic mean $F_1$ are applied in information retrieval tasks \citep{RefWorks:249}.\\
For each label $\lambda_j:j=1\ldots q$, the summation of true positives (${tp}_j$), true negatives (${tn}_j$), false positives (${fp}_j$), and false negatives (${fn}_j$) are calculated according to the classifier applied. Then, the binary performance evaluation metric \textit{B} can be calculated with either macro or micro-averaging operations:
\begin{equation}
    Macro\ B=\frac{1}{q}\sum_{j=1}^{q}{B({tp}_j,{tn}_j,{fp}_j,{fn}_j)}
\end{equation}
\begin{equation}
    Micro\ B=B\left(\sum_{j=1}^{q}{tp}_j,\sum_{j=1}^{q}{tn}_j,\sum_{j=1}^{q}{fp}_j,\sum_{j=1}^{q}{fn}_j\right)
\end{equation}
Therefore, the definitions of \textit{Precision} (\textit{P}), \textit{Recall} (\textit{R}), and $F_1$ are easily derived as \citep{RefWorks:245}:
\begin{equation}
    Micro\ Precision\ (MiP)=\frac{\sum_{j=1}^{q}{tp}_j}{\sum_{j=1}^{q}{tp}_j+\sum_{j=1}^{q}{fp}_j}	
\end{equation}
\begin{equation}
    Micro\ Recall\ (MiR)=\frac{\sum_{j=1}^{q}{tp}_j}{\sum_{j=1}^{q}{tp}_j+\sum_{j=1}^{q}{fn}_j}	
\end{equation}
\begin{equation}
    Micro\ F_1=\frac{2\times MiR\times MiP}{MiR+MiP}
\end{equation}
\begin{equation}
    Macro\ Precision=\frac{1}{q}\sum_{j=1}^{q}\frac{{tp}_j}{{tp}_j+{fp}_j}	
\end{equation}
\begin{equation}
    Macro\ Recall=\frac{1}{q}\sum_{j=1}^{q}\frac{{tp}_j}{{tp}_j+{fn}_j}	
\end{equation}
\begin{equation}
    Macro\ F_1=\frac{1}{q}\sum_{j=1}^{q}\frac{2\times R_j\times P_j}{R_j{+P}_j}; R_j=\frac{{tp}_j}{{tp}_j+{fn}_j}, P_j=\frac{{tp}_j}{{tp}_j+{fp}_j}
\end{equation}
where \textit{Macro} $F_1$ is the harmonic mean of \textit{Precision} and \textit{Recall} based on first averaging each label $\lambda_j$ and then averaging over all labels. On the other hand, \textit{Micro} $F_1$ is the harmonic mean of \textit{Micro Precision} and \textit{Micro Recall} as defined above.

For all of the above metrics, they score on a scale of zero to one, where a higher score implies better alignment between the predicted label set and the ground truth set.
\subsection{Semantic Similarity}
The current formulations of the above metrics share a significant drawback as they consistently overlook the inherent semantic similarity between each label. For example, let’s assume the ground truth multi-label set is \small\{"bicycle," "child," "helmet," "road," "tree"\small\}, and the predicted set is \small\{"bike," "boy," "trail," "tree," "grass," "flower"\small\}. Evaluating the similarity between the two label sets with the above metrics will consider only the label “tree” as a true positive and overlook the close semantic similarity between the labels \small\{“child,” “boy”\small\}, \small\{“bicycle,” “bike”\small\} and \small\{“road,” “trail”\small\}. 

To overcome this misrepresentation, a straightforward adjustment can be made. For each of the above example-based metrics (\textit{Accuracy, Precision, Recall,} and $F_1$), the correct predictions can be decided not by the exact predicted label, but rather the semantic similarity between the predicted and true labels. Here, we use the cosine similarity (${p\cdot r}/{\Vert p \Vert \Vert r \Vert}$) between the  word2vec embeddings \citep{RefWorks:307} of the predicted (\textit{p}) and real (\textit{r}) labels, where the correct prediction is considered above certain threshold.\footnote{Here, the threshold is set to \textit{0.4}.}

Additionally, we applied the word mover’s distance (WMD) metric \citep{RefWorks:315}, which is an earth mover’s distance based method  \citep{RefWorks:319, RefWorks:320}, and aimed at evaluating the semantic distance between two documents. Let us denote $Y_i^\ast={y_{i,j}:j=1,\ldots,r}$ as the ground truth label set of image \textit{i}, and $Z_i^\ast={z_{i,s}:s=1,\ldots,p}$ as the label set of image \textit{i} predicted by the multi-label classifier \textit{h}, $Z_i^\ast=h\left(x_i\right)$. Note that $Y_i^*$ and $Z_i^*$ include the explicit label set (e.g., \small\{"bike," "boy," "trail," "tree," "grass," "flower"\small\}), where \textit{r} and \textit{p} don’t have to be on the same size. Defining the two label sets as two bag-of-words (BOW) allows us to apply the WDM method to evaluate their semantic distance. The WDM algorithm requires that the two BOW are represented as a normalized BOW (nBOW) vector $d\in\mathbb{R}^n$, where $n=r\cup p$, and $d_l=t_l/\sum_{k=1}^{n}t_k: t_l$ is the number of times that the word \textit{l} of \textit{n} appears in the BOW. Let \textit{d} be the nBOW representation of $Y_i^*$ and $d^\prime$ of $Z_i^*$. The second requirement of the WDM is a semantic distance evaluation between every two labels, where \textit{c(l,k)} is referred to as the cost of “traveling” from word \textit{l} to word \textit{k}. Therefore, Let $W\in\mathbb{R}^{dim\times n}$ be the word2vec embedding matrix, where $w_k\in\mathbb{R}^{dim}$ is the dim-dimensional embedding representation of word \textit{k} from the vocabulary of \textit{n} words. Hence, the “traveling cost” from word \textit{l} to word \textit{k} is defined as their Euclidean distance, $c(l,k)=||w_l-w_k||$. Next, let us define a sparse flow matrix ${T\in\mathbb{R}}^{n\times n}$, where $t_{l,k}\geq0$ represents the ratio of participation of word \textit{l} from \textit{d} to travel to word \textit{k} from $d^\prime$. It is clear that a word can participate in traveling as much as its nBOW $t_l$ ratio, therefore the $\sum_{k} t_{l,k}=d_l$ and $\sum_{l} t_{l,k}=d_k^\prime$ participation ratio restrictions are applied. Finally, the distance between the two BOW can be defined as the minimum sum of the weighted traveling cost from \textit{d} to $d^\prime$
\begin{equation}
    wdm=min\sum_{l,k=1}^{n}{t_{l,k}c\left(l,k\right)}
\end{equation}
subject to the participation ratio restrictions. Since the \textit{wmd} calculate the minimum traveling distance a score of 0 is considered as a perfect match. For our purposes, we average the \textit{wmd}s for all of the \textit{n} images in the tested dataset for every API:
\begin{equation}
    WMD=\frac{1}{n}\sum_{i=1}^{n}{wdm_i}
\end{equation}

The above metrics are based on a single word embedding to evaluate the labels' semantic similarity; while the semantic example-based metrics use the embeddings within known metrics, the WMD takes it one step further, and considers the aggregated similarity between the ground-truth and predicted BOWs. As aggregated understanding can be beneficial for semantic similarity \citep{Kubany2020TripletDivergence}, we propose to leverage the aggregated embeddings of the BOWs as a means to find their aggregated similarity.\footnote{Calculated by the cosine similarity.} The BERT \citep{Devlin2018Bert:Understanding}, RoBERTa \citep{Liu2019Roberta:Approach}, and XLNet \citep{Yang2019Xlnet:Understanding} are bidirectional transformer-based methods and considered as the state-of-the-art approaches to embed an entire text to a single embedding \citep{Wolf2019HuggingFacesProcessing}. We also consider the semantic similarity fine-tuned versions of BERT and RoBERTa\footnote{Here we used the 'bert-base-nli-stsb-wkpooling' and 'roberta-base-nli-stsb-mean-tokens' pretrained versions.} methods as they demonstrate superior performance in semantic similarity tasks \citep{Reimers2019Sentence-BERT:BERT-Networks}.

\section{Results and Discussion}
\subsection{Experiment Setup}
\textbf{Testing Dataset:} The testing dataset should include images from multiple domains, as well as multiple semantic annotations of objects, concepts, or labels, to ensure as close to real-life evaluation as possible. Some of the commercial APIs apply limits regarding the number of image requests for multi-label classification during a period of time and in total. Given these limitations, we evaluated the APIs’ performance with the first 1,000 images\footnote{Sorted in name ascending order. We selected the first 1,000 images that have objects, as some of the images do not have them.} from each dataset, which, to our understanding, are sufficient for satisfactory performance evaluation of the examined APIs. 
\begin{itemize}
    \item \textbf{Visual Genome dataset:} The Visual Genome (VG) dataset\footnote{We used the 1.0 version of the dataset.} \citep{Krishna2017VisualAnnotations} consists of 108,077 everyday multi-domain images, which represent the intersection between the MS-COCO \citep{RefWorks:202} and the YFCC100M \citep{Thomee2016YFCC100M:Research} datasets. Each image in the dataset is associated with an average of 21 objects (out of 75,729 possibilities) for multi-label classification purposes. Within the 1000 images subset, there are 3728 possible objects, an average of 14.1 objects per image, and 1.05 labels per object, where 8.5\% of the used labels are unknown to the word2vec embedding model.
    \item \textbf{Open Images dataset:} The Open Images (OI) dataset\footnote{We used the sixth version of the dataset.} \citep{Krasin2017OpenImages:Classification., Kuznetsova2020TheScale} incorporate $\sim$9M images of diverse sceneries collected from the "Flickr" online service. For multi-label classification, each image includes an average of 8.3 objects out of 600 classes. Within the 1000 subset, each image includes an average of 3.9 objects out of 263 object classes, where 16.5\% of the 3.9 average objects are unknown to the word2vec embedding model.
\end{itemize}
It is essential to consider the applied training data for the various APIs. We assume that the commercial APIs vendors continuously attempt to improve their services. Since the commercial APIs’ training data is unknown, selecting widely accessible and popular datasets makes it more likely to be considered, and therefore should confine the predicament of biased evaluation. For the open-source APIs, we chose the ImageNet pre-trained models, which is well-known simple scenery dataset and should provide as an adequate baseline.\\
\textbf{APIs' Evaluated Objects:} Some of the commercial APIs restrict the number of predicted objects per image, while others predict only a few object labels with high confidence and low confidence for others. Evaluating with different top levels allow a fair comparison between the APIs. In this study, we perform the APIs evaluation based on three object levels: the top five, three, and one label(s) according to their confidence level (see Tables \ref{tbl:top 1 label VG}-\ref{tbl:top 5 labels OI}). Also, for a fair comparison, we queried all the APIs using their vanilla versions without any specific fine-tuning.

\begin{table}[htbp]
  \centering
  \resizebox{\columnwidth}{!}{%
    \begin{tabular}{cccccccc}
    \toprule
    \textbf{API} & \textbf{Type} & \textbf{Training} &   \textbf{Unknown Labels} & \textbf{Mean} \\
    & & \textbf{Data} &  \textbf{out of Top Five} & \textbf{Labels} \\
    &  &  & \textbf{Objects (\%)} & \textbf{Per Object} \\
    &  &  & \textbf{(VG / OI)} & \textbf{(VG / OI)} \\
    \midrule
    
    Clarifai & Commercial & unknown &   0.5 / 6.5 & 1 / 1 \\
    Google Cloud Vision & Commercial & unknown &   10 / 15.4 & 1 / 1\\
    IBM Watson & Commercial & unknown &    0.8 / 8.1 & 1 / 1\\
    Immaga & Commercial & unknown &  5.6 / 5.7 & 1 / 1 \\
    Microsoft Computer Vision & Commercial & unknown &   1.1 / 2.7 & 1 / 1\\
    Wolfram & Commercial & unknown &   42.3 / 23.8 & 1 / 1\\
    DeepDetect & Open Source & ImageNet &    10.4 / 11 & 1.95 / 1.89 \\
    InceptionReNet-v2 & Open Source & ImageNet &  11.9 / 11.1 & 1.98 / 1.95 \\
    Inception-v3 & Open Source & ImageNet &  12.1 / 11.4 & 1.97 / 1.94 \\
    MobileNet-v2 & Open Source & ImageNet &  11.4 / 10.6 & 1.97 / 1.93\\
    ResNet50 & Open Source &  ImageNet &  10.3 / 9.2 & 1.99 / 1.95\\
    ResNet50 & Open Source & COCO &  5.2 / 3.9 & 1 / 1\\
    VGG19 & Open Source & ImageNet &  9.6 / 10.2 & 2 / 1.93\\
    YOLO-v3 & Open Source & ImageNet & 19.6 / 16 & 1 / 1\\
    YOLO-v3 & Open Source & COCO & 5.5 / 4 & 1 / 1\\

    \bottomrule
    \end{tabular}%
   }
    \caption{APIs’ metadata.}
    \label{tbl:APIs settings}
\end{table}

\subsection{Example-Based Metrics}
One of the first observations is that in general, the examined APIs have relatively low scores. A few factors can explain this observation; first, there is the issue of model settings and training data, and although we apply well-known datasets to reduce testing bias, we do not know which training data was used by the commercial APIs, on the other hand, the open-source APIs behave as expected as they all were pre-trained with the simple ImageNet dataset. Nevertheless, since the commercial APIs achieve higher scores than the open-source APIs, it might suggest that at least some images of the datasets were in their training data. Additionally, although the commercial APIs' out of the box configurations also contribute to the low scores, they are necessary if we wish for a fair comparison without prior knowledge of its structure. Hence, we analyze the commercial and open-source APIs separately.

Second, the VG dataset holds an average of 14.1 objects per image, while we account for a maximum of five predictions, this explains the low scores of the \textit{Accuracy}, \textit{Recall}, and $F_1$ metrics, as they consider the number of the ground truth objects' labels. On the other hand, the OI dataset holds an average of 3.9 objects per image, thus, provide a larger scale of \textit{Accuracy}, \textit{Recall}, and $F_1$ metrics' scores, with the same scale for the \textit{Precision} metric.\\
\textbf{Commercial APIs:} The commercial APIs' performance is consistent on both datasets, and reveal that four APIs stand out with high scores: Microsoft Computer Vision (MCV), Imagga, IBM and Google APIs consistently hold top places, with the MCV API dramatically outperform others (inhabit the most of green cells in Tables \ref{tbl:top 1 label VG}-\ref{tbl:top 5 labels OI}). It is worth noting that the Google API performance is not consistent between the datasets, as it holds a higher place in the OI dataset. We might explain this top performance to the fact that Google also manufactures the OI dataset, and it could have been used in its training. Having a high \textit{Precision} score means that most of the predictions made by the MCV API are relevant, with only a few false positives. It also has a high \textit{Recall} scores, indicating it predicted relatively more of the ground truth labels (with only a few false negatives); this is also reflected in the relatively high \textit{Accuracy} score. The MCV's top $F_1$ score also reassures its dominance as it is the harmonic mean of its high scores in the \textit{Precision} and \textit{Recall} metrics. Since the number of true labels is higher than the predicted ones, in our view, the \textit{Precision} metric gives a more reliable indication of the APIs performance. Here, the MCV's \textit{Precision} dramatically outperforms others, which means that its order of predictions is closer to the true labels than other APIs.

\begin{sidewaysfigure}[ht]
    \centering
    \includegraphics[width=\paperwidth]{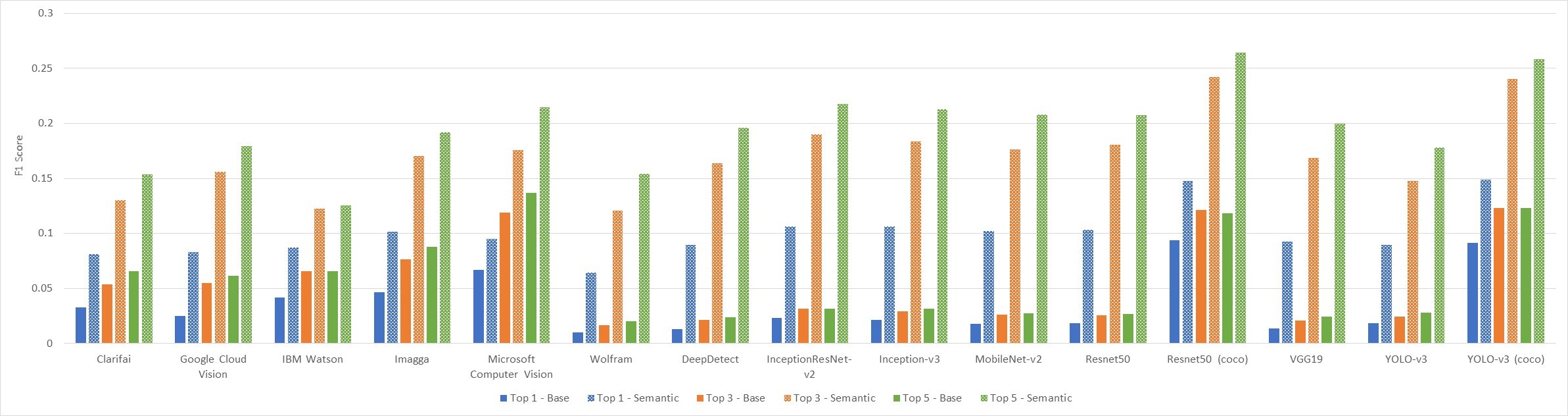}
    \caption{The $F_1$ Results for each prediction level of the evaluated APIs on the Visual Genome dataset (higher scores are better).}
    \label{fig: F1 VG}

    \centering
    \includegraphics[width=\paperwidth]{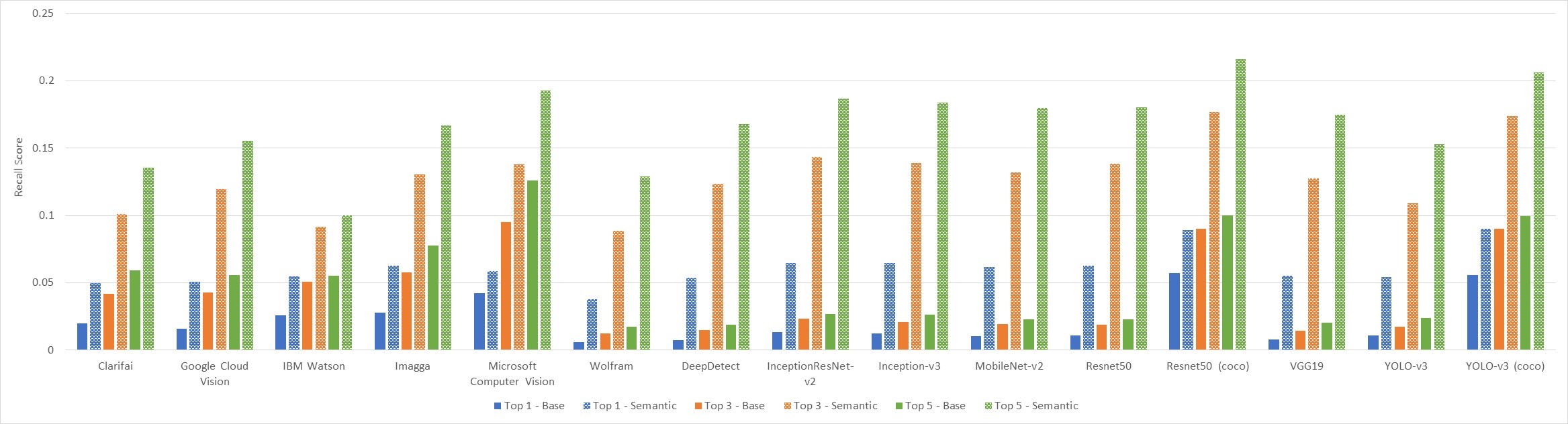}
    \caption{The \textit{Recall} Results for each prediction level of the evaluated APIs on the Visual Genome dataset (higher scores are better).}
    \label{fig: recall VG}
\end{sidewaysfigure}

\begin{sidewaysfigure}[ht]
        \centering
    \includegraphics[width=\paperwidth]{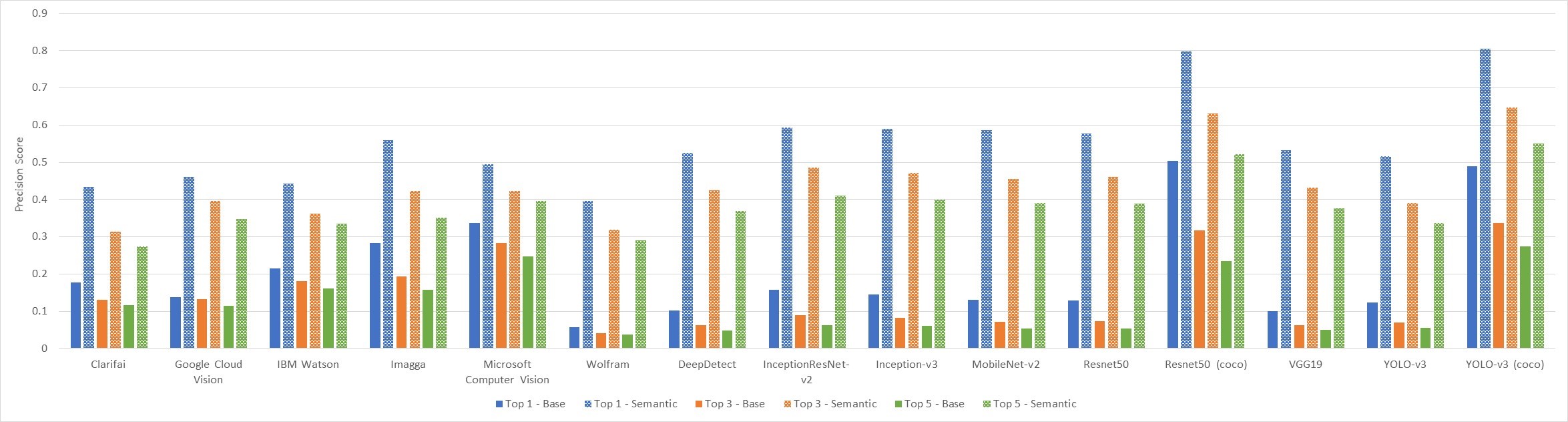}
    \caption{The \textit{Precision} Results for each prediction level of the evaluated APIs on the Visual Genome dataset (higher scores are better).}
    \label{fig: precision VG}

    \centering
    \includegraphics[width=\paperwidth]{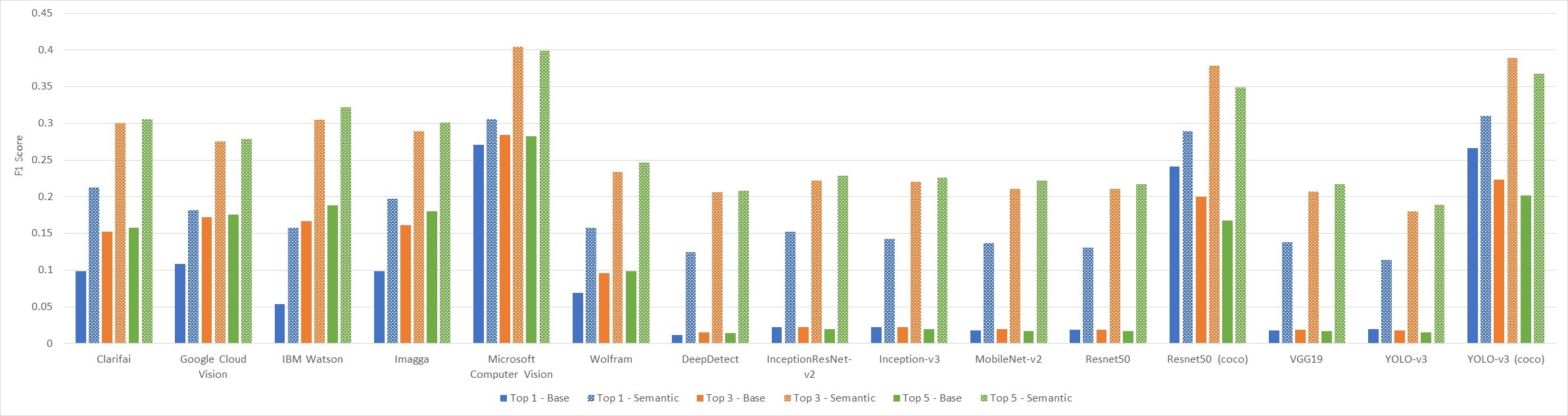}
    \caption{The $F_1$ Results for each prediction level of the evaluated APIs on the Open Images dataset (higher scores are better).}
    \label{fig: F1 OI}
\end{sidewaysfigure}

\begin{sidewaysfigure}[ht]

    \centering
    \includegraphics[width=\paperwidth]{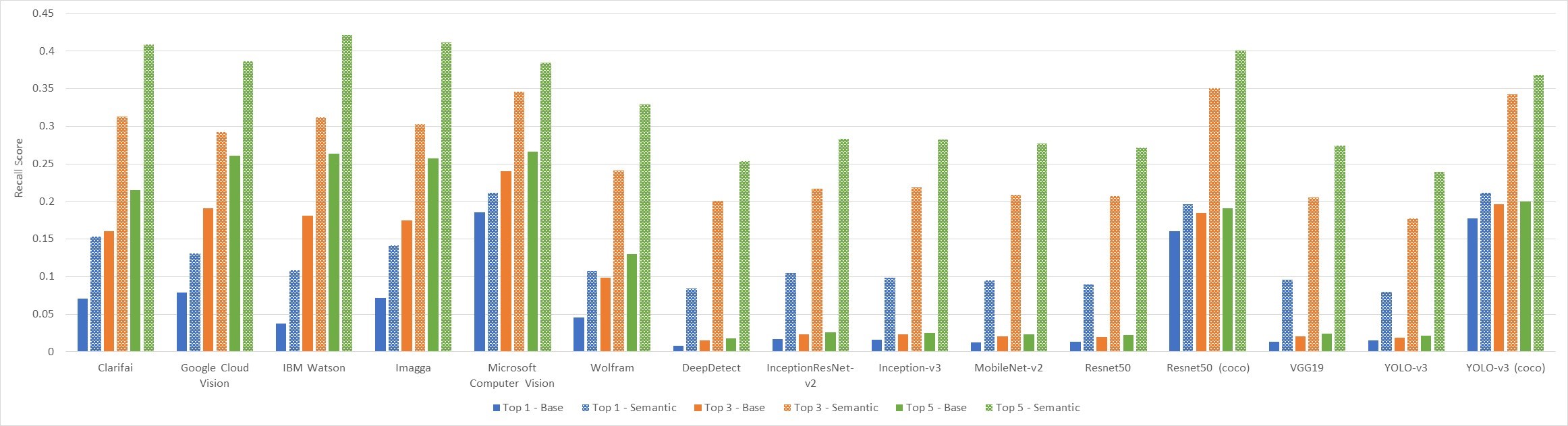}
    \caption{The \textit{Recall} Results for each prediction level of the evaluated APIs on the Open Images dataset (higher scores are better).}
    \label{fig: recall OI}

        \centering
    \includegraphics[width=\paperwidth]{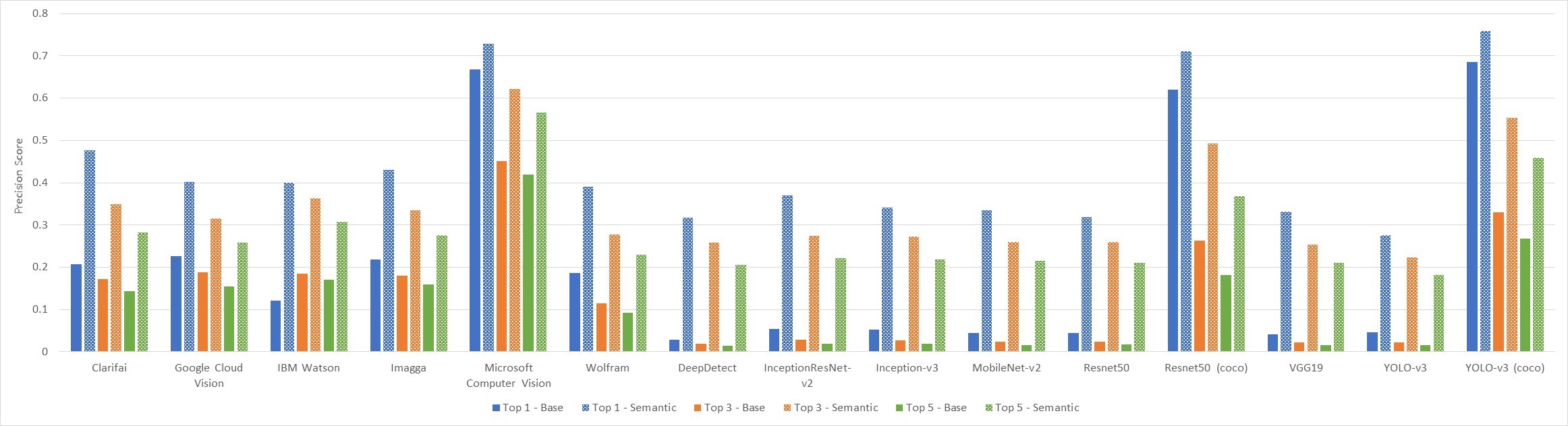}
    \caption{The \textit{Precision} Results for each prediction level of the evaluated APIs on the Open Images dataset (higher scores are better).}
    \label{fig: precision OI}
\end{sidewaysfigure}

Another perspective is the score dynamic between the different top predicted levels. As expected, as the number of predicted labels rise, the \textit{Precision} scores decrease (Figures \ref{fig: precision VG} and \ref{fig: precision OI}) as it is more likely to be correct in one label than in five, and the \textit{Recall} scores increase (Figures \ref{fig: recall VG} and \ref{fig: recall OI}) as it more likely to find more labels in common with the true labels. The increasing $F_1$ scores (Figures \ref{fig: F1 VG} and \ref{fig: F1 OI}) teach us that the \textit{Recall} dynamic change is stronger than the \textit{Precision} one. 
\\
\textbf{Open-Source APIs:} As expected, the open-source APIs produce much lower results than the commercial APIs. These relatively low results can be explained by their training data, as they are all trained on the ImageNet dataset, which includes much simpler sceneries and different labels than the datasets. With that saying, the APIs consisted of more elaborated network architecture yield better performance, usually in relation to their specific engineering advances, with the general performance order of InceptionResNet-v2, Inception-v3, MobileNet-v2, ResNet50, YOLO-v3, and VGG19. As before, we particularly notice the performance differences within the \textit{Precision} metric, which demonstrate the InceptionResNet-v2 and Inception-v3 dominance. We will revisit the analysis of these APIs within the semantic analysis (section \ref{sec: semantic metrics}).

The superior performance of the commercial APIs, and in particular, of the MCV API raises the question of whether their top performance is due to their network structure, training data, or both. We can only partially answer this question since the complete details of commercial APIs is unknown; fortunately, we know that the top-performing MCV API is based on the ResNet architecture \citep{RefWorks:338}, but unsure of its specific network details and training data. Therefore, to confine this question, we compare its performance on the VG dataset with the ResNet50 and YOLO-v3 APIs trained on both ImageNet and COCO\footnote{The VG dataset in the intersection between the MS-COCO and the YFCC100M datasets.} datasets (Table \ref{tbl:microsoft reference VG}). We can see that the performance change of the ResNet50 and YOLO-v3 APIs between the ImageNet and the COCO pre-trained models are consistent in scale. Let us analyze the MCV API performance under the assumption that Microsoft would offer the best possible network architecture they have in their arsenal for their payable API. Considering that, if the MCV would have been trained only on the COCO dataset, it needed to outperform the COCO pre-trained ResNet50, as it is the leanest flavor of ResNet with 50 layers, where deeper and better-performing networks with 101 and 152 layers exist \citep{RefWorks:338}. Since, in general, the MCV yield scores higher than than the ImageNet pre-trained ResNet50 and lower than the COCO pre-trained ResNet50, in particular within the top one and three predictions, we are left to conclude that it is not trained solely on the COCO dataset, and can remain in the evaluation. Also, we perform the same analysis of the MCV, ResNet50, and YOLO-v3 performance on the OI dataset (Table \ref{tbl:microsoft reference OI}). Since, as far as we know, the OI does not include the COCO dataset, and the MCV API dominant over the ResNet50 COCO API, it reassures our previous conclusion that the MCV API is trained on more images than those inhabit the COCO dataset.\\ There is an immense performance jump when training with the COCO dataset for the ResNet50 and YOLO-v3 APIs on both datasets. The performance turnover is so significant for the YOLO-v3 API on the OI dataset, that it changes its rank from the last place to among top performers, especially in the top predicted label. This performance change highlights the drawback of training with a simple dataset as the ImageNet and makes us wonder about the potential performance improvement of the other open-source APIs, which demonstrate better performance than YOLO-v3 when trained only on the ImageNet dataset.

From the example-based metrics perspective, the MCV is the top all-around performer; nevertheless, if other APIs are needed, the Imagga and IBM are excellent options.

\subsection{Label-Based Metrics}
\begin{figure}
    \centering
      \begin{subfigure}[b]{0.49\textwidth}
    \includegraphics[width=\textwidth]{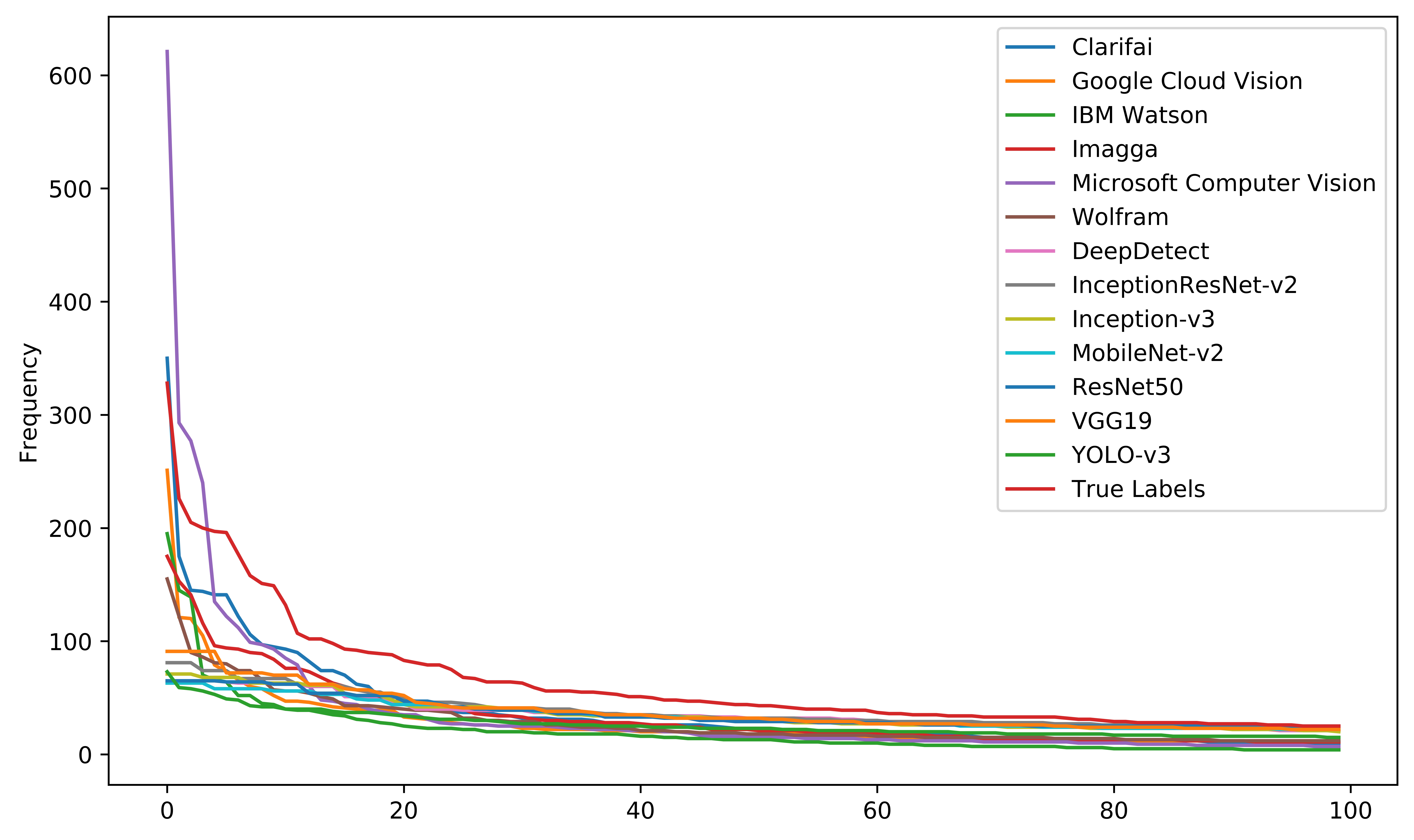}
    \caption{Visual Genome.}
    \label{fig: label freq VG}
  \end{subfigure}
  \begin{subfigure}[b]{0.49\textwidth}
    \includegraphics[width=\textwidth]{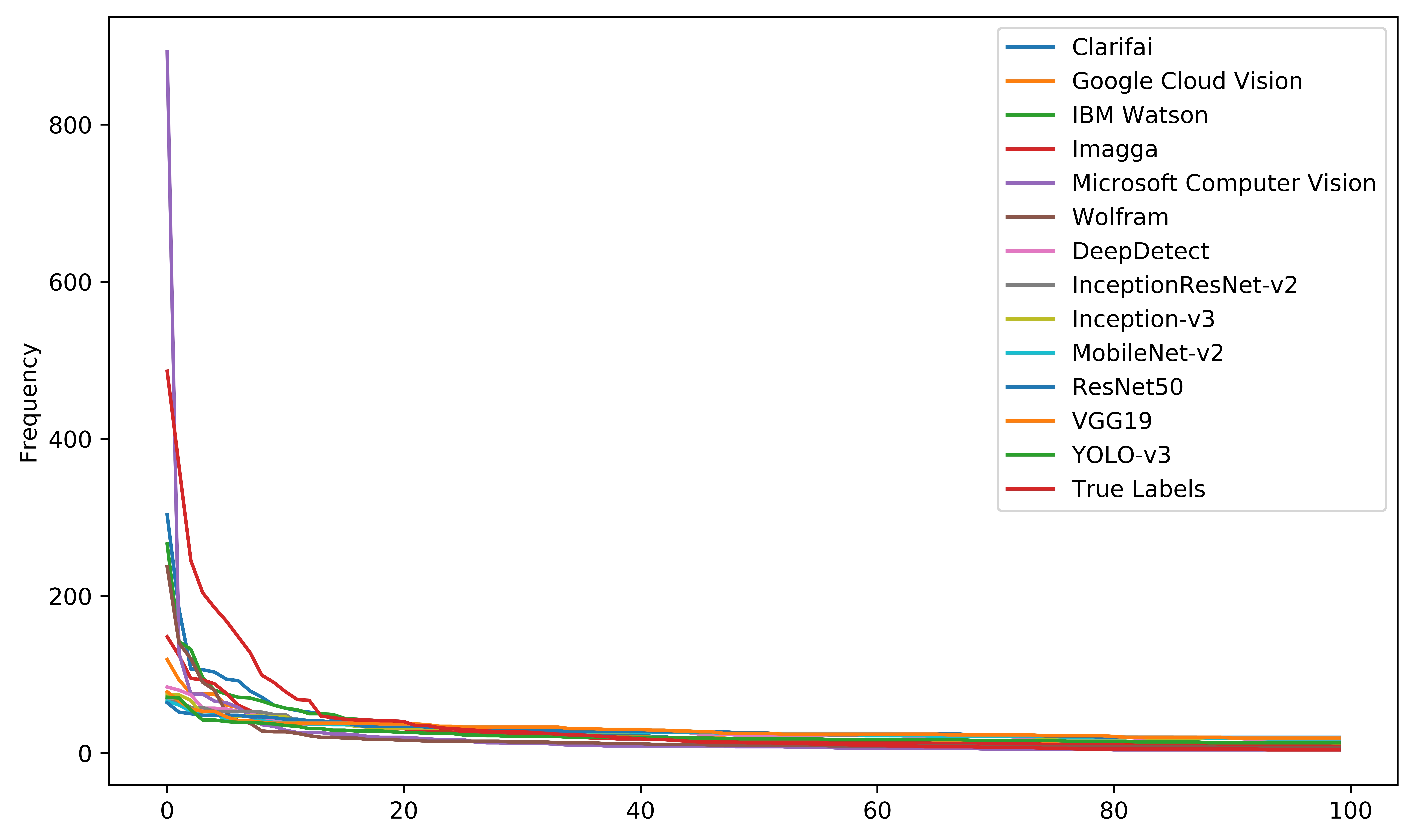}
    \caption{Open Images.}
    \label{fig: label freq OI}
  \end{subfigure}
    \caption{100 most Frequent object classes.}
    \label{fig: label freq}
\end{figure}

Within this type of metrics, we evaluate the performance of the various APIs from the label perspective (see Tables \ref{tbl:top 1 label VG}-\ref{tbl:top 5 labels OI}). In the \textit{Macro} family of metrics, we evaluate the performance of predicting each label separately and then averaging them all, whereas, in the \textit{Micro} metrics, we evaluate the performance of all the labels' predictions together. Furthermore, since the \textit{Macro} metrics does not account for the false predictions (\textit{fp} for \textit{Precision} and \textit{fn} for \textit{Recall}) when the true positive is zero, and the \textit{Micro} metrics does, we can evaluate the prediction balance between the labels. We notice low \textit{Macro} scores for the VG dataset, whereas the OI dataset allows much higher scores. The VG low scores suggest that many labels have zero true positives, which agree with a higher number of object classes (\textit{3728} in the VG, and \textit{263} in the OI) and a long tail object class distribution (Figure \ref{fig: label freq}). The score difference of the \textit{Recall} metric in between the VG and OI datasets, continue to support this line of thinking. The same score scale of the \textit{Recall Micro} and \textit{Recall Macro} metrics, with even a slightly higher \textit{Recall Micro} score in the VG dataset, further indicates that the zero true positives are of infrequent labels. On the other hand, while the score scale of the \textit{Precision} metrics in the OI dataset is about the same, in the VG dataset, it is not. As before, The VG's low \textit{Precision Macro} score is due to the many zero true positives, while the higher \textit{Precision Micro} score indicates that there are less false positives. Still, the \textit{Precision Micro} score in the VG dataset is lower than in the OI dataset due to the division in dataset's object class number.
\\
\textbf{Commercial APIs:} Like with the \textit{example-based} metrics, the MCV, IBM, and Imagga APIs, stand out on both datasets. However, the Google Cloud Vision APIs, which demonstrate only occasional good performance on the VG dataset, outperform all APIs in the OI dataset. As before, we suspect that the exceptional top scores of Google Cloud Vision API on the OI dataset might suggest that the Google published dataset is part of its training set.\\
\textbf{Open-Source APIs:} For these APIs, besides within the \textit{Micro Precision} metric, the performing differences between the best and worst APIs are marginal, making it very hard to gain any knowledge. Nevertheless, there are two points worthy of pointing out; first, within the \textit{Macro Recall} metric, for the first time, although in a small margin, the InceptionResNet-v2 and the Inception-v3 APIs consistently outperform all others in the VG dataset, meaning that on average they are slightly more capable of predicting the correct label. Second, for the \textit{Micro Precision} metric, the YOLO-v3 API performs better than other open-source APIs on the VG dataset, following by the InceptionResNet-v2, Inception-v3, and ResNet50 APIs.

The MCV, Imagga, IBM, and Google APIs are ahead with the MCV outperforming all other considering the scores from both the example and label-based metrics. Additionally, the InceptionResNet-v2 and the Inception-v3 APIs demonstrate some good performance, and it would be wise to give them further consideration.

\begin{figure}
    \centering
    \includegraphics[scale=0.26]{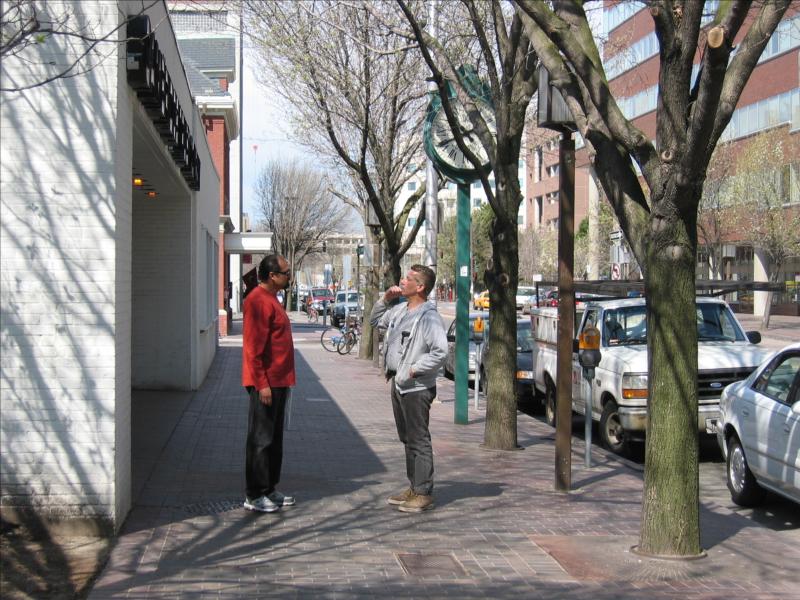}
    \caption{Example image '1.jpg' from the Visual Genome dataset}
    \label{fig: image 1 @ VG}
\end{figure}

\begin{table}
  \centering
  \resizebox{\textwidth}{!}{%
    \begin{tabular}{llccccccc}
    \toprule
     API & Labels & Recall & Recall & Precision & Precision & WMD & Fine-Tuned  & Fine-Tuned 
      \\
       &  &  &  (Semantic) &  &  (Semantic) &  &  BERT &  RoBERTa \\
    \midrule
    Clarifai &
      city, \underline{vehicle}, people, road, \textbf{street} &
      0.04 & 0.08 & 0.2 & 0.4 & 3.19 & 0.28 & {0.13}
      \\
    Google Cloud Vision &
      city, \underline{pedestrian}, \textbf{street}, signage, walking &
      {0.04} & {0.08} & {0.2} & {0.4} & {3.48} & {0.27} & {0.19}
      \\
    IBM Watson &
      \textbf{street}, city, crowd, people &
      {0.04} & {0.04} & {0.25} & {0.25} & {3.35} & {0.13} & {-0.02}
      \\
    Imagga &
      \textbf{sidewalk}, walk, \textbf{street}, road, city &
      {0.08} & {0.08} & {0.4} & {0.4} & {3.30} & {0.29} & {0.20}
      \\
    Microsoft Computer Vision &
     outdoor, \textbf{building}, \textbf{street}, road, \textbf{sidewalk} &
      {0.12} & {0.12} & {0.6} & {0.6} & {3.20} & {0.28} & {0.29} 
      \\
    Wolfram &
      road, path, container, conveyance, \underline{vehicle} &
      {0} & {0.04} & {0} & {0.2} & {3.62} & {0.25} & {0.02}
      \\
    DeepDetect &
      ski, crutch, [prison, prison house], sliding door, shovel  &
      {0} & {0} & {0} & {0} & {3.69} & {0.25} & {0.21} \\
      
    InceptionResNet-v2 &
      \textbf{parking meter}, [traffic light, traffic signal, \underline{stoplight}],   &
      {0.04} & {0.08} & {0.2} & {0.4} & {3.80} & {0.35} & {0.25} \\
      & [pay-phone, pay-station], [mailbox, letter box], & & & & & & & \\
      & [cash machine, cash dispenser, automated teller machine] & & & & & & & \\
    Inception-v3 &
      [jinrikisha, ricksha, \underline{rickshaw}],    &
      {0} & {0.04} & {0} & {0.2} & {3.77} & {0.48} & {0.30} \\
      & [ashcan, trash can, garbage can, wastebin], & & & & & & & \\
      & [streetcar, tram,tramcar, trolley, trolley car], & & & & & & & \\
      & [bookshop, bookstore, bookstall], plastic bag & & & & & & & \\
    MobileNet-v2 &
      \textbf{parking meter}, [jinrikisha, ricksha, \underline{rickshaw}],  &
      {0.04} & {0.12} & {0.2} & {0.6} & {3.57} & {0.45} & {0.26} \\
      & [police van, police wagon, paddy wagon], & & & & & & & \\
      & [\underline{cab}, hack, taxi, taxicab], crutch & & & & & & & \\

    Resnet50 &
      \textbf{parking meter}, [mailbox, letter box],   &
      {0.04} & {0.12} & {0.2} & {0.6} & {3.69} & {0.47} & {0.29} \\
      & [ashcan, trash can, garbage can, wastebin], & & & & & & & \\
      & [traffic light, traffic signal, \underline{stoplight}],  & & & & & & & \\
      & [jinrikisha, ricksha, \underline{rickshaw}] & & & & & & & \\
    Resnet50 (COCO) &
      \underline{person}, \textbf{car}, \underline{bicycle}, traffic light, truck &
      {0.04} & {0.12} & {0.2} & {0.6} & {3.20} & {0.35} & {0.24} \\
    VGG19 &
      \textbf{parking meter}, [jinrikisha, ricksha, \underline{rickshaw}],  &
      {0.04} & {0.08} & {0.2} & {0.4} & {3.78} & {0.34} & {0.25} \\
      & [gas pump, gasoline pump, petrol pump], ski, ambulance & & & & & & & \\

    YOLO-v3 &
      pole, cash machine, guillotine, plastic bag, \underline{jean} &
      {0} & {0.02} & {0} & {0.2} & {3.83} & {0.26} & {0.07} \\
    YOLO-v3 (COCO) &
      \underline{person}, \textbf{car}, truck, \underline{bicycle}, \textbf{parking meter} &
      {0.08} & {0.16} & {0.4} & {0.8} & {3.26} & {0.36} & {0.17} \\
    \midrule
    Ground Truth &
      arm, back, bike, bikes, building, car, chin, clock, glasses, guy,  & & & & & & &\\
      & headlight, jacket, lamp post, man, pants, parking meter, shade,  & & & & & & & \\
      & shirt, shoes, sidewalk, sign, sneakers, street, tree, tree trunk & & & & & & & \\
    \bottomrule
    \end{tabular}
    }
  \caption{The APIs' top five labels for image '1.jpg' from the Visual Genome dataset (Figure \ref{fig: image 1 @ VG}). The table's annotations refer to \textbf{bold} labels as \textit{tp} and the \underline{underline} labels as semantic similar \textit{tp}.}

  \label{tbl:APIs' results for image 1 @ VG}
\end{table}

\subsection{Semantic Metrics}\label{sec: semantic metrics}
To demonstrate the importance of semantic metrics within the multi-label evaluation, we refer to the APIs' top five predicted labels of image 1 from the VG dataset as an example (Figure \ref{fig: image 1 @ VG}, Table \ref{tbl:APIs' results for image 1 @ VG}). Within this image, let us take the first API as an example, in human prospective the Clarifai API includes many valid labels, but only the \textit{"street"} label is correct according to the ground-truth labels list. For instance, the \textit{"vehicle"} label is not included in the ground-truth label set, and traditionally not considered as a true positive prediction. Obviously, it has the same meaning as the ground-truth \textit{"car"} label, and a similarity score of \textit{0.78}, and therefore, is considered as semantic true positive, as it should be. Since we are more interested in the semantic meaning of the labels rather than their exact wording, this example, like many others, showcase the necessity of semantic metrics for multi-label evaluation, in particular when the APIs are trained on a different object class list from the examined dataset.

The word2vec embedding model is well-known to be an accurate and comprehensive word representation. Nevertheless, despite our efforts,\footnote{For example, the label "parking meter" for image 1, which is first lowercased and cleaned from unwanted chars, does not exist in the word2vec model. We try different permutation of the label: without space ("parkingmeter"), with an underscore ("parking\_meter"), with first caps and underscore ("Parking\_Meter"). In this case, the "Parking\_Meter" label permutation exists in the word2vec model.} some of the APIs' predicted labels are not found within the model and considered as the "unknown" label (see Table \ref{tbl:APIs settings} for APIs' "unknown" rates). We further discuss the effect of these settings with regard to semantic example-based and WMD metrics. \\
\textbf{Semantic Example-Based Metrics:} Evaluating the APIs' performance with the classic approach stated the MCV, IBM, and Imagga APIs as top performers with the MCV outperforming all, the semantic evaluation on the VG dataset shed a new light of the APIs performance. In this evaluation, the MCV is still a high performer, but now the InceptionResNet-v2, Inception-v3, and frequently the ResNet50 and MobileNet-v2 APIs demonstrate top performance, meaning they can predict more closely semantically related labels. These findings are even more dramatic as they are trained on the ImageNet dataset and have about ten percent of "unknown" labels (Table \ref{tbl:APIs settings}), even if considering that they predict about twice the labels per object, their rate of unknowns is still higher. A higher percentage of "unknowns" makes it harder to recognize a predicted label as true positive, which lowers the performance metrics score. These findings demonstrate the performance advantage of a more elaborated network structure over simpler and shallower networks (like the DeepDetect API); moreover, it highlights the importance of the residual ideas as it exists in three of the five top semantic performers (InceptionResNet-v2, MCV, and ResNet50). The semantic evaluation on the OI dataset incline to less dramatic results for the open-source APIs in terms of top performers. Nevertheless, the simple dataset trained APIs, which naturally achieve much lower scores than the commercial APIs with the traditional metrics, now measure on the same semantic score scale. We relate the lower open-source semantic scores in the OI dataset to its lower average objects per image (14.1 in the VG vs. 3.9 in the OI), as the semantic metrics benefit from more labels to be potentially semantically similar. These results further demonstrate the benefits of the semantic metrics, which can compare APIs with a lesser effect of their training dataset.\\
It is important to note that these findings of the performance dominance of the more elaborated APIs (InceptionResNet-v2, Inception-v3, and frequently the ResNet50) found here are consistent with prior ImageNet evaluations \citep{Silberman2016TensorFlow-SlimLibrary}, and further validate our findings.\\
Additionally, the high "unknown" rates of the Wolfram (42.3\% / 23.8\%), and YOLO-v3 (19.6\% / 16\%) APIs can partially explain their semantic low scores. When such a significant portion of the labels predicted as "unknown" and different from the actual labels, it makes no surprise they have such low scores.\\
\textbf{WMD Metric:} Following the superior performance of the InceptionResNet-v2, Inception-v3, ResNet50 APIs in the semantic example-based metrics, we are somewhat surprised as to their lower scores with the WMD metric. Nevertheless, in our view, a simple explanation exists. If we take the top five level case as an example, each of the open-source APIs predicts about ten labels per image (two labels per object, see Table \ref{tbl:APIs settings}), except for the YOLO-v3 API which predicts five labels. Considering their "unknown" rates, on average, all of the open-source APIs include one "unknown" label, which is much rarer in the commercial APIs. As the WMD metric calculates the minimum traveling distance between the true and predicted nBOW label vectors, the inclusion of the "unknown" label in the nBOW is forcing the semantic distance to be much higher, hence lowering the open-source APIs scores.\\
\textbf{Labels' BOW Embedding:} Considering a BOW aggregated embedding allows us to overcome the issue of "unknown" labels and to estimate the aggregated labels' semantic similarity directly. Within this evaluation, the vanilla version of the BERT, RoBERTa, and XLNet methods produced poor results with $\mathtt{\sim}0.007$ standard deviation between the APIs scores. These results are consistent with previous findings in which their vanilla versions are prone to poor sentence embeddings \citep{Reimers2019Sentence-BERT:BERT-Networks}. In contrast, the semantic similarity scores of the fine-tuned versions of the BERT and RoBERTa methods\footnote{We have not found the semantic similarity fine-tuned version of XLNet.} demonstrate conclusive results on the VG dataset. These findings demonstrate that the results of both embeddings agree with each other, and strongly supports the previous semantic example-based metrics results, in which the InceptionResNet-v2, Inception-v3, and ResNet50 APIs consistently demonstrate top performance. As with the \textit{Semantic Example-Based} metrics, the open-source APIs perform better with VG dataset than with the OI dataset. The BERT and RoBERTa methods benefit from more input words to produce more accurate embeddings (up to a point) and the lesser amount of the OI objects per image, in particular in the face of a large amount of BOW predicted labels of the open-source APIs harm their semantic similarity score. Therefore, the performance evaluation of the open-source APIs with the OI dataset is less informative than with the VG dataset. Within this context, it is essential to remember that the issue of input word number is critical for accurate embeddings for every use in such methods.

Whereas the non-semantic metrics exhibit the dominance of the MCV, Imagga, and IBM APIs, The semantic metrics challenge their dominance and allow the simpler dataset trained APIs to be considered as equals, and even weigh the InceptionResNet-v2, Inception-v3, and ResNet50 APIs as the top semantic performers.

\section{Conclusions}
In this study, we compared the performance of some of the most prominent deep learning multi-label classification APIs. Throughout our evaluations using the traditional metrics approaches, the MCV, IBM, and Imagga APIs consistently demonstrate top performance with the MCV API as the top performer; obviously, their performance is no match for the open-source APIs which are trained with a much simpler dataset. However, the semantic metrics allow these low starting point APIs to be evaluated as equals and ever consider the InceptionResNet-v2, Inception-v3, and ResNet50 APIs among the top semantic performers. These evaluations demonstrate the capabilities and added value of the semantic metrics in obtaining profound insights regarding the labels meaning even when training with a simple dataset and a different object-class dictionary, insights, which are unavailable otherwise.

As the field of multi-label classification advances, we believe that the proposed semantic metrics and the performance comparison performed in this study can be beneficial for both researchers and users in the quest for image understanding.
\section*{Acknowledgments}
This study was supported by grants from the MAGNET program of the Israeli Innovation Authority and the MAFAT program of the Israeli Ministry of Defense.

{\small

\bibliography{references_mendeley}
}


\begin{landscape}
\begin{table}
  \centering
  \resizebox{\paperwidth}{!}{%
\begin{tabular}{ccccccccc|cccccc|ccc}
    \toprule
     \textbf{API}& \textbf{Accuracy } & \textbf{Accuracy} & \textbf{Recall} & \textbf{Recall} & \textbf{Precision} & \textbf{Precision} & \textbf{F1} &
     \textbf{F1} & \textbf{Macro} & \textbf{Macro} & \textbf{Macro} & \textbf{Micro} & \textbf{Micro} & \textbf{Micro} & \textbf{WMD} & \textbf{Fine-Tuned} & \textbf{Fine-Tuned}\\
      & & \textbf{(Semantic)} &  & \textbf{(Semantic)} & & \textbf{(Semantic)} & & \textbf{(Semantic)}  & \textbf{Precision} & \textbf{Recall} & \textbf{F1} & \textbf{Precision} & \textbf{Recall} & \textbf{F1}  & 
      & \textbf{BERT} & \textbf{RoBERTa}\\
    \midrule
    Clarifai &
      \cellcolor[rgb]{ .851,  .878,  .51}0.020 &
      \cellcolor[rgb]{ .988,  .761,  .486}0.050 &
      \cellcolor[rgb]{ .851,  .878,  .51}0.020 &
      \cellcolor[rgb]{ .988,  .761,  .486}0.050 &
      \cellcolor[rgb]{ .878,  .89,  .514}0.178 &
      \cellcolor[rgb]{ .98,  .557,  .447}0.434 &
      \cellcolor[rgb]{ .847,  .878,  .51}0.033 &
      \cellcolor[rgb]{ .988,  .714,  .475}0.081 &
      \cellcolor[rgb]{ 1,  .922,  .518}0.007 &
      \cellcolor[rgb]{ .98,  .592,  .455}0.002 &
      \cellcolor[rgb]{ .98,  .624,  .459}0.002 &
      \cellcolor[rgb]{ .992,  .788,  .49}0.201 &
      \cellcolor[rgb]{ .878,  .886,  .514}0.011 &
      \cellcolor[rgb]{ .878,  .89,  .514}0.020 &
      \cellcolor[rgb]{ .502,  .776,  .486}3.658 &
      \cellcolor[rgb]{ .973,  .427,  .42}0.162 &
      \cellcolor[rgb]{ .973,  .412,  .42}0.106
      \\
\bottomrule
    \end{tabular}
    }
     \caption{The APIs' metrics results with the top predicted label per image evaluated on the Visual Genome dataset. The color gradient scales the APIs performance from green (best) to red (worst).}
\label{tbl:top 1 label VG}
\end{table}

\begin{table}[htbp]
  \centering
\resizebox{\paperwidth}{!}{%
    \begin{tabular}{ccccccccc|cccccc|ccc}
    \toprule
   \textbf{API}& \textbf{Accuracy } & \textbf{Accuracy} & \textbf{Recall} & \textbf{Recall} & \textbf{Precision} & \textbf{Precision} & \textbf{F1} &
     \textbf{F1} & \textbf{Macro} & \textbf{Macro} & \textbf{Macro} & \textbf{Micro} & \textbf{Micro} & \textbf{Micro} & \textbf{WMD} & \textbf{Fine-Tuned} & \textbf{Fine-Tuned}\\
      & & \textbf{(Semantic)} &  & \textbf{(Semantic)} & & \textbf{(Semantic)} & & \textbf{(Semantic)}  & \textbf{Precision} & \textbf{Recall} & \textbf{F1} & \textbf{Precision} & \textbf{Recall} & \textbf{F1}  & 
      & \textbf{BERT} & \textbf{RoBERTa}\\
      \midrule
    Clarifai &
      \cellcolor[rgb]{ .839,  .875,  .51}0.030 &
      \cellcolor[rgb]{ .976,  .514,  .439}0.077 &
      \cellcolor[rgb]{ .831,  .875,  .51}0.042 &
      \cellcolor[rgb]{ .98,  .573,  .451}0.101 &
      \cellcolor[rgb]{ .851,  .882,  .51}0.131 &
      \cellcolor[rgb]{ .973,  .412,  .42}0.314 &
      \cellcolor[rgb]{ .835,  .875,  .51}0.054 &
      \cellcolor[rgb]{ .976,  .51,  .435}0.130 &
      \cellcolor[rgb]{ .545,  .792,  .494}0.012 &
      \cellcolor[rgb]{ .988,  .706,  .475}0.006 &
      \cellcolor[rgb]{ .992,  .796,  .49}0.005 &
      \cellcolor[rgb]{ 1,  .922,  .518}0.160 &
      \cellcolor[rgb]{ .843,  .878,  .51}0.024 &
      \cellcolor[rgb]{ .847,  .878,  .51}0.042 &
      \cellcolor[rgb]{ .714,  .839,  .498}3.508 &
      \cellcolor[rgb]{ .973,  .443,  .424}0.253 &
      \cellcolor[rgb]{ .973,  .412,  .42}0.220
      \\
\bottomrule
    \end{tabular}%
    }
        \caption{The APIs' metrics results with the top three predicted labels per image evaluated on the Visual Genome dataset. The color gradient scales the APIs performance from green (best) to red (worst).}
        \label{tbl:top 3 labels VG}
\end{table}%

\begin{table}[htbp]
  \centering
   \resizebox{\paperwidth}{!}{%
    \begin{tabular}{ccccccccc|cccccc|ccc}
    \toprule
 \textbf{API}& \textbf{Accuracy } & \textbf{Accuracy} & \textbf{Recall} & \textbf{Recall} & \textbf{Precision} & \textbf{Precision} & \textbf{F1} &
     \textbf{F1} & \textbf{Macro} & \textbf{Macro} & \textbf{Macro} & \textbf{Micro} & \textbf{Micro} & \textbf{Micro} & \textbf{WMD} & \textbf{Fine-Tuned} & \textbf{Fine-Tuned}\\
      & & \textbf{(Semantic)} &  & \textbf{(Semantic)} & & \textbf{(Semantic)} & & \textbf{(Semantic)}  & \textbf{Precision} & \textbf{Recall} & \textbf{F1} & \textbf{Precision} & \textbf{Recall} & \textbf{F1}  & 
      & \textbf{BERT} & \textbf{RoBERTa}\\
      \midrule
    Clarifai &
      \cellcolor[rgb]{ .812,  .871,  .51}0.037 &
      \cellcolor[rgb]{ .98,  .584,  .451}0.089 &
      \cellcolor[rgb]{ .8,  .867,  .51}0.059 &
      \cellcolor[rgb]{ .984,  .678,  .471}0.136 &
      \cellcolor[rgb]{ .824,  .871,  .51}0.116 &
      \cellcolor[rgb]{ .973,  .412,  .42}0.274 &
      \cellcolor[rgb]{ .804,  .867,  .51}0.066 &
      \cellcolor[rgb]{ .98,  .612,  .455}0.153 &
      \cellcolor[rgb]{ .545,  .792,  .494}0.015 &
      \cellcolor[rgb]{ .988,  .733,  .478}0.009 &
      \cellcolor[rgb]{ .843,  .878,  .51}0.007 &
      \cellcolor[rgb]{ .906,  .894,  .514}0.149 &
      \cellcolor[rgb]{ .792,  .863,  .506}0.036 &
      \cellcolor[rgb]{ .8,  .867,  .51}0.057 &
      \cellcolor[rgb]{ .725,  .843,  .502}3.445 &
      \cellcolor[rgb]{ .98,  .573,  .447}0.297 &
      \cellcolor[rgb]{ .98,  .557,  .447}0.273
      \\
\bottomrule
    \end{tabular}%
    }
   \caption{The APIs' metrics results with the top five predicted labels per image evaluated on the Visual Genome dataset. The color gradient scales the APIs performance from green (best) to red (worst).}
   \label{tbl:top 5 labels VG}
\end{table}%

\begin{table}
  \centering
 \resizebox{\paperwidth}{!}{%
    \begin{tabular}{cccccccccc|cccccc|ccc}
    \toprule
   \textbf{API}& \textbf{Top} & \textbf{Accuracy } & \textbf{Accuracy} & \textbf{Recall} & \textbf{Recall} & \textbf{Precision} & \textbf{Precision} & \textbf{F1} &
     \textbf{F1} & \textbf{Macro} & \textbf{Macro} & \textbf{Macro} & \textbf{Micro} & \textbf{Micro} & \textbf{Micro} & \textbf{WMD} & \textbf{Fine-Tuned} & \textbf{Fine-Tuned}\\
      && & \textbf{(Semantic)} &  & \textbf{(Semantic)} & & \textbf{(Semantic)} & & \textbf{(Semantic)}  & \textbf{Precision} & \textbf{Recall} & \textbf{F1} & \textbf{Precision} & \textbf{Recall} & \textbf{F1}  & 
      &\textbf{BERT} & \textbf{RoBERTa}\\
    \midrule
         Microsoft Computer Vision &
      1 &
      \cellcolor[rgb]{ 1,  .922,  .518}0.042 &
      \cellcolor[rgb]{ .984,  .694,  .471}0.059 &
      \cellcolor[rgb]{ 1,  .922,  .518}0.042 &
      \cellcolor[rgb]{ .984,  .694,  .471}0.059 &
      \cellcolor[rgb]{ 1,  .922,  .518}0.337 &
      \cellcolor[rgb]{ .973,  .412,  .42}0.495 &
      \cellcolor[rgb]{ 1,  .922,  .518}0.067 &
      \cellcolor[rgb]{ .98,  .624,  .459}0.095 &
      \cellcolor[rgb]{ .388,  .745,  .482}0.011 &
      \cellcolor[rgb]{ .973,  .412,  .42}0.002 &
      \cellcolor[rgb]{ .973,  .412,  .42}0.0034 &
      \cellcolor[rgb]{ 1,  .922,  .518}0.3240 &
      \cellcolor[rgb]{ 1,  .922,  .518}0.020 &
      \cellcolor[rgb]{ 1,  .922,  .518}0.038 &
      \cellcolor[rgb]{ 1,  .922,  .518}3.614 &
      \cellcolor[rgb]{ .973,  .412,  .42}0.195 &
      \cellcolor[rgb]{ .973,  .412,  .42}0.111
      \\
\bottomrule
    \end{tabular}%
    }
     \caption{Results for the Microsoft Computer Vision, ResNet50 and YOLO-v3 APIs trained on the ImageNet and COCO dataset and evaluated on the Visual Genome dataset. The color gradient scales the APIs performance from green (best) to red (worst).}
   \label{tbl:microsoft reference VG}
\end{table}%

\begin{table}
  \centering
  \resizebox{\paperwidth}{!}{%
    \begin{tabular}{ccccccccc|cccccc|ccc}
    \toprule
     \textbf{API}& \textbf{Accuracy } & \textbf{Accuracy} & \textbf{Recall} & \textbf{Recall} & \textbf{Precision} & \textbf{Precision} & \textbf{F1} &
     \textbf{F1} & \textbf{Macro} & \textbf{Macro} & \textbf{Macro} & \textbf{Micro} & \textbf{Micro} & \textbf{Micro} & \textbf{WMD} & \textbf{Fine-Tuned} & \textbf{Fine-Tuned}\\
      & & \textbf{(Semantic)} &  & \textbf{(Semantic)} & & \textbf{(Semantic)} & & \textbf{(Semantic)}  & \textbf{Precision} & \textbf{Recall} & \textbf{F1} & \textbf{Precision} & \textbf{Recall} & \textbf{F1}  & 
      & \textbf{BERT} & \textbf{RoBERTa}\\
    \midrule
           Clarifai &
      \cellcolor[rgb]{ .804,  .867,  .51}0.071 &
      \cellcolor[rgb]{ .725,  .843,  .502}0.153 &
      \cellcolor[rgb]{ .804,  .867,  .51}0.071 &
      \cellcolor[rgb]{ .725,  .843,  .502}0.153 &
      \cellcolor[rgb]{ .851,  .878,  .51}0.207 &
      \cellcolor[rgb]{ .82,  .871,  .51}0.476 &
      \cellcolor[rgb]{ .816,  .871,  .51}0.098 &
      \cellcolor[rgb]{ .761,  .855,  .506}0.212 &
      \cellcolor[rgb]{ .761,  .855,  .506}0.166 &
      \cellcolor[rgb]{ .988,  .761,  .486}0.075 &
      \cellcolor[rgb]{ .871,  .886,  .514}0.091 &
      \cellcolor[rgb]{ .443,  .761,  .486}0.863 &
      \cellcolor[rgb]{ .835,  .875,  .51}0.053 &
      \cellcolor[rgb]{ .816,  .871,  .51}0.099 &
      \cellcolor[rgb]{ .698,  .831,  .498}3.241 &
      \cellcolor[rgb]{ .757,  .851,  .506}0.315 &
      \cellcolor[rgb]{ .71,  .839,  .502}0.290
      \\
\bottomrule
    \end{tabular}
    }
     \caption{The APIs' metrics results with the top predicted label per image evaluated on the Open Images dataset. The color gradient scales the APIs performance from green (best) to red (worst).}
\label{tbl:top 1 label OI}
\end{table}

\begin{table}[htbp]
  \centering
\resizebox{\paperwidth}{!}{%
    \begin{tabular}{ccccccccc|cccccc|ccc}
    \toprule
   \textbf{API}& \textbf{Accuracy } & \textbf{Accuracy} & \textbf{Recall} & \textbf{Recall} & \textbf{Precision} & \textbf{Precision} & \textbf{F1} &
     \textbf{F1} & \textbf{Macro} & \textbf{Macro} & \textbf{Macro} & \textbf{Micro} & \textbf{Micro} & \textbf{Micro} & \textbf{WMD} & \textbf{Fine-Tuned} & \textbf{Fine-Tuned}\\
      & & \textbf{(Semantic)} &  & \textbf{(Semantic)} & & \textbf{(Semantic)} & & \textbf{(Semantic)}  & \textbf{Precision} & \textbf{Recall} & \textbf{F1} & \textbf{Precision} & \textbf{Recall} & \textbf{F1}  & 
      & \textbf{BERT} & \textbf{RoBERTa}\\
      \midrule
        Clarifai &
      \cellcolor[rgb]{ .706,  .839,  .502}0.098 &
      \cellcolor[rgb]{ .761,  .855,  .506}0.202 &
      \cellcolor[rgb]{ .616,  .812,  .498}0.160 &
      \cellcolor[rgb]{ .545,  .792,  .494}0.313 &
      \cellcolor[rgb]{ .796,  .863,  .506}0.172 &
      \cellcolor[rgb]{ .871,  .886,  .514}0.350 &
      \cellcolor[rgb]{ .698,  .835,  .502}0.152 &
      \cellcolor[rgb]{ .741,  .847,  .506}0.300 &
      \cellcolor[rgb]{ .678,  .831,  .502}0.253 &
      \cellcolor[rgb]{ .878,  .886,  .514}0.157 &
      \cellcolor[rgb]{ .737,  .847,  .506}0.172 &
      \cellcolor[rgb]{ .467,  .769,  .49}0.783 &
      \cellcolor[rgb]{ .631,  .816,  .498}0.131 &
      \cellcolor[rgb]{ .6,  .808,  .498}0.225 &
      \cellcolor[rgb]{ .635,  .816,  .494}3.014 &
      \cellcolor[rgb]{ .675,  .827,  .502}0.383 &
      \cellcolor[rgb]{ .675,  .827,  .502}0.342
      \\
\bottomrule
    \end{tabular}%
    }
        \caption{The APIs' metrics results with the top three predicted labels per image evaluated on the Open Images dataset. The color gradient scales the APIs performance from green (best) to red (worst).}
        \label{tbl:top 3 labels OI}
\end{table}%

\begin{table}
  \centering
   \resizebox{\paperwidth}{!}{%
    \begin{tabular}{ccccccccc|cccccc|ccc}
    \toprule
 \textbf{API}& \textbf{Accuracy } & \textbf{Accuracy} & \textbf{Recall} & \textbf{Recall} & \textbf{Precision} & \textbf{Precision} & \textbf{F1} &
     \textbf{F1} & \textbf{Macro} & \textbf{Macro} & \textbf{Macro} & \textbf{Micro} & \textbf{Micro} & \textbf{Micro} & \textbf{WMD} & \textbf{Fine-Tuned} & \textbf{Fine-Tuned}\\
      & & \textbf{(Semantic)} &  & \textbf{(Semantic)} & & \textbf{(Semantic)} & & \textbf{(Semantic)}  & \textbf{Precision} & \textbf{Recall} & \textbf{F1} & \textbf{Precision} & \textbf{Recall} & \textbf{F1}  & 
      & \textbf{BERT} & \textbf{RoBERTa}\\
      \midrule
         Clarifai &
      \cellcolor[rgb]{ .706,  .839,  .502}0.096 &
      \cellcolor[rgb]{ .769,  .855,  .506}0.198 &
      \cellcolor[rgb]{ .522,  .784,  .49}0.215 &
      \cellcolor[rgb]{ .447,  .765,  .486}0.409 &
      \cellcolor[rgb]{ .812,  .871,  .51}0.143 &
      \cellcolor[rgb]{ .894,  .89,  .514}0.283 &
      \cellcolor[rgb]{ .678,  .831,  .502}0.158 &
      \cellcolor[rgb]{ .725,  .843,  .502}0.305 &
      \cellcolor[rgb]{ .71,  .839,  .502}0.268 &
      \cellcolor[rgb]{ .906,  .894,  .514}0.189 &
      \cellcolor[rgb]{ .741,  .847,  .506}0.199 &
      \cellcolor[rgb]{ .478,  .773,  .49}0.729 &
      \cellcolor[rgb]{ .533,  .788,  .494}0.182 &
      \cellcolor[rgb]{ .506,  .78,  .49}0.291 &
      \cellcolor[rgb]{ .651,  .82,  .494}3.018 &
      \cellcolor[rgb]{ .663,  .827,  .502}0.391 &
      \cellcolor[rgb]{ .671,  .827,  .502}0.348
      \\
\bottomrule
    \end{tabular}%
    }
   \caption{The APIs' metrics results with the top five predicted labels per image evaluated on the Open Images dataset. The color gradient scales the APIs performance from green (best) to red (worst).}
   \label{tbl:top 5 labels OI}
\end{table}%

\begin{table}
  \centering
 \resizebox{\paperwidth}{!}{%
    \begin{tabular}{cccccccccc|cccccc|ccc}
    \toprule
   \textbf{API}& \textbf{Top} & \textbf{Accuracy } & \textbf{Accuracy} & \textbf{Recall} & \textbf{Recall} & \textbf{Precision} & \textbf{Precision} & \textbf{F1} &
     \textbf{F1} & \textbf{Macro} & \textbf{Macro} & \textbf{Macro} & \textbf{Micro} & \textbf{Micro} & \textbf{Micro} & \textbf{WMD} & \textbf{Fine-Tuned} & \textbf{Fine-Tuned}\\
      && & \textbf{(Semantic)} &  & \textbf{(Semantic)} & & \textbf{(Semantic)} & & \textbf{(Semantic)}  & \textbf{Precision} & \textbf{Recall} & \textbf{F1} & \textbf{Precision} & \textbf{Recall} & \textbf{F1}  & 
      &\textbf{BERT} & \textbf{RoBERTa}\\
    \midrule
    Microsoft Computer Vision &
      1 &
      \cellcolor[rgb]{ .388,  .745,  .482}0.185 &
      \cellcolor[rgb]{ .392,  .749,  .486}0.212 &
      \cellcolor[rgb]{ .388,  .745,  .482}0.185 &
      \cellcolor[rgb]{ .392,  .749,  .486}0.212 &
      \cellcolor[rgb]{ .557,  .796,  .494}0.667 &
      \cellcolor[rgb]{ .784,  .859,  .506}0.729 &
      \cellcolor[rgb]{ .388,  .745,  .482}0.271 &
      \cellcolor[rgb]{ .514,  .784,  .49}0.306 &
      \cellcolor[rgb]{ .388,  .745,  .482}0.242 &
      \cellcolor[rgb]{ .388,  .745,  .482}0.116 &
      \cellcolor[rgb]{ .388,  .745,  .482}0.1417 &
      \cellcolor[rgb]{ .388,  .745,  .482}0.8674 &
      \cellcolor[rgb]{ .561,  .796,  .494}0.156 &
      \cellcolor[rgb]{ .388,  .745,  .482}0.264 &
      \cellcolor[rgb]{ .388,  .745,  .482}2.795 &
      \cellcolor[rgb]{ .388,  .745,  .482}0.420 &
      \cellcolor[rgb]{ .388,  .745,  .482}0.359
      \\
\bottomrule
    \end{tabular}%
    }
     \caption{Results for the Microsoft Computer Vision, ResNet50 and YOLO-v3 APIs trained on the ImageNet and COCO dataset and evaluated on the Open Images dataset. The color gradient scales the APIs performance from green (best) to red (worst).}
   \label{tbl:microsoft reference OI}
\end{table}%

\end{landscape}

\end{document}